**Kyriaki IOANNIDOU**
**Programme Interdisciplinaire d'Études Approfondies en Interprétation et Traduction**
**Laboratoire de Traduction et de Traitement Automatique du Langage, Université Aristote de Thessalonique, Grèce**
kiroanni@auth.gr

**Elsa TOLONE**
**Laboratoire d'informatique Gaspard-Monge, Université Paris-Est, France**
**& Facultad de Matemática, Astronomía y Física, Universidad Nacional de Córdoba, Argentine**
elsa.tolone@univ-paris-est.fr

# CONSTRUCTION DU LEXIQUE *LGLEX* A PARTIR DES TABLES DU LEXIQUE-GRAMMAIRE DES VERBES DU GREC MODERNE

**Résumé**
Dans cet article, nous dressons un bilan du travail effectué sur les ressources du grec moderne concernant le Lexique-Grammaire des verbes[1]. Nous détaillons les propriétés définitoires de chaque table, ainsi que l'ensemble des changements effectués sur les intitulés de propriétés afin de les rendre cohérents. Grâce à l'élaboration de la table des classes, regroupant l'ensemble des propriétés, nous avons pu envisager la conversion des tables en un lexique syntaxique : *LGLex*. Ce lexique, au format texte ou XML, est généré par l'outil *LGExtract* (Constant & Tolone, 2010). C'est un format directement exploitable dans les applications de Traitement Automatique des Langues (TAL).
Mots clés : Lexique-Grammaire, verbes du grec moderne, table des classes, *LGLex,* lexique syntaxique

## 1. Introduction

Depuis une dizaine d'années, un programme de description formalisée de la langue grecque pour l'analyse syntaxico-sémantique a permis de construire des tables du Lexique-Grammaire, en adoptant la méthodologie proposée par Gross (1975), Boons, Guillet et Leclère (1976a ; 1976b). À l'heure actuelle, nous disposons de 17 tables de verbes simples (Yannacopoulou, 2005 ; Fista, 2007 ; Kyriacopoulou, 2010 ; Voskaki, 2011), 7 de noms prédicatifs (Sfetsiou, 2007), 16 tables d'adverbes (semi-)figés (Voyatzi, 2006 ; Voyatzi & Kakoyianni-Doa, 2010) et 2 tables de noms composés (Kyriakopoulou, 2011)[2]. Dans le cadre de notre recherche, nous avons pris en compte les tables des prédicats verbaux, le même travail étant en cours pour les noms prédicatifs.

Afin de procéder à l'analyse syntaxique des textes grecs, il faut intégrer les données du Lexique-Grammaire dans un analyseur syntaxique, ce qui nécessite la conversion des données dans un format exploitable en Traitement Automatique des Langues (TAL). Étant fortement inspirés des travaux déjà réalisés pour le français (Tolone, 2011), nous avons suivi la démarche suivante pour les prédicats verbaux :
− collecter toutes les tables ;
− créer une table des classes, afin d'y faire figurer les propriétés définitoires décrites dans la littérature et de rendre cohérent l'ensemble des tables, comme cela a été fait pour le français (Tolone, 2009 ; Tolone *et al.*, 2010) ;
− créer un script interprétant chaque intitulé de la table des classes qui, exécuté par l'outil *LGExtract* (Constant & Tolone, 2010), permet de convertir les tables au format *LGLex*.

[2] Ces tables seront disponibles ultérieurement sur http://www-igm.univ-mlv.fr/~ressourcesgrec.

Dans cet article, après la présentation des tables utilisées pour la construction du lexique syntaxique, nous allons lister les types de modifications effectuées lors de la mise en cohérence des propriétés. Par la suite, nous allons évoquer la création de la table des classes détaillée dans (Ioannidou & Kyriacopoulou, 2010) qui nous a permis d'ajouter les propriétés définitoires et d'assurer l'homogénéisation des intitulés des propriétés. Enfin, nous allons présenter le lexique syntaxique obtenu.

## 2. Ressources linguistiques et propriétés définitoires des tables

Les tables verbales ainsi que les propriétés définitoires vraies pour l'ensemble des entrées de chaque table sont les suivantes (la première construction étant la construction de base)[3] :

| | |
|---|---|
| 32GA : N0 V N1 | avec N1 apparition |
| 32GD : N0 V N1 | avec N1 disparition |
| 32GC : N0 V N1 | avec N1 =: Nconc |
| 32GCL : N0 V N1 | avec N1 =: Npc |
| 32GCV : N0 V N1 | avec N0 Vsup N1 Prep V-n |
| 32GH : N0 V N1 | avec N1 =: Nhum |
| 32GNM : N0 V N1 | avec N1 =: N-hum |
| 32GPL : N0 V N1 | avec N1 =: Npl obl |
| 32GRA : N0 V N1 | avec N0 Vsup N1 V-adjaccusatif *[équivalent à V-adj]*[4] |
| 4G : N0 V N1 | avec une complétive en N0 |
| 6G : N0 V N1 | avec une complétive en N1 |
| 9G : N0 V N1 σε *[se=à]* N2 | avec une complétive en N1 |
| 38GL : N0 V N1 Loc N2 source Loc N3 destination | |
| 38GLS : N0 V N1 Loc N2 source | |
| 38GLD : N0 V N1 Loc N2 destination | |
| 38GLH : N0 V N1 Loc N2 destination | avec N1 =: Nhum |
| 38GLR : N0 V N1 Loc N2 | |

L'ensemble des tables verbales représentent 3 966 emplois verbaux décrits. Notons que pour les verbes, certaines lignes peuvent donner lieu à plusieurs entrées puisque des préfixes sont également codés, ce qui totalise 4 252 entrées en comptant les verbes préfixés. Certes, ces tables sont loin de couvrir toute la langue grecque mais la constitution de cette ressource est en cours. Le nombre de ressources en Grèce n'étant pas très important, il s'agit de la plus grande ressource syntaxique à ce jour.

## 3. Modifications dans les tables

Pour procéder à la conversion du contenu de plusieurs tables en un seul fichier, il faut que l'ensemble des propriétés respecte les mêmes conventions de notation et de structure dans les différentes tables. De ce fait, nous avons effectué des modifications dans les tables à l'aide de la table des classes (voir section 4). Pour effectuer ces modifications, nous nous sommes basées sur la documentation des propriétés des verbes du français[5]. Cependant, nous avons pris en compte les particularités de la langue grecque, en essayant d'être le plus proche possible des notations françaises. Par exemple, l'encodage particulier qu'exige la langue grecque nous a conduit à ne pas garder les accents français dans les notations. Nous avons adopté donc les mêmes symboles en enlevant les accents (*Prép* → *Prep* pour préposition[6]).

Les changements effectués concernent toutes les colonnes des tables, sauf celles des exemples et des traductions qui n'ont pas encore été traitées. Ils peuvent être regroupés en cinq catégories : les modifications typographiques, les modifications au niveau de la structure des intitulés, l'ajout des informations lexicales, la suppression des colonnes qui contenaient des propriétés définitoires ou non pertinentes et les changements purement linguistiques.

---
[3] Les notations entre *[ ]* indiquent la romanisation du mot grec, suivi de sa traduction en français. Elles ont été ajoutées pour des questions de lisibilités mais ne figurent pas dans les propriétés.
[4] Pour le grec, les cas peuvent être spécifiés à droite d'un complément.
[5] Disponible avec les tables du Lexique-Grammaire du français sur le site http://infolingu.univ-mlv.fr/ > Données Linguistiques > Lexique-Grammaire > Téléchargement.
[6] Dans cet article, le changement va être noté sous la forme *x* ⟶ *y* où *x* est la notation avant le changement et *y* la notation d'après.

1. Parmi les erreurs typographiques, on rencontre des fautes dues à l'alternance de caractères grecs et latins, ou de caractères majuscules et minuscules (*ppv* → *Ppv*). D'autres modifications au niveau typographique concernent la présence ou non d'espaces (*N0=:Nhum* → *N0 =: Nhum*), le tronquage ou non d'une notation (*disp* → *disparition*), ou l'utilisation d'une notation différente (*V-ος [-os]* → *V-adj, Sfx = ος [os]*) (Ioannidou & Kyriacopoulou, 2010). Nous avons, de plus, utilisé la notation *x-V* où *x* correspond à un préfixe qui est ajouté aux verbes grecs (p.ex. *εκ-V [ek-]*, *συν-V [syn-]*, etc.) (Fista, 2007). Cette notation a une interprétation particulière lors de la conversion des tables au format *LGLex*. Si cette colonne vaut + pour une entrée verbale donnée, il faut spécifier cette nouvelle entrée qui correspond au verbe préfixé partageant les mêmes propriétés que le verbe non préfixé. Les changements au niveau typographique représentent la plus grande partie des changements effectués dans les tables (55%).

2. À part les modifications qui concernent les symboles utilisés dans les intitulés des propriétés, nous avons effectué des changements qui concernent la structure des propriétés (utilisation de la virgule ou des symboles =:, =, utilisation d'une structure différente pour désigner un trait sémantique et un rôle thématique). Pour les notations qui existent seulement pour le grec, nous avons utilisé des structures proches de celles adoptées pour les autres intitulés, pour faciliter leur conversion au format *LGLex*. Par exemple, au lieu d'utiliser la notation *Pfxεκ-[ek-]/source* qu'on utilisait pour exprimer l'ajout du préfixe *εκ* qui donne la notion de source, nous avons noté la construction complète *N0 εκ-V [ek-] N1 Loc N2 source* qui est conforme aux conventions du Lexique-Grammaire. Dans le tableau ci-dessous, nous avons regroupé les différentes structures utilisées pour les intitulés dans les tables grecques[7]. Les modifications qui concernent la structure des notations représentent 30% des changements effectués.

|    | Type d'information | Exemple |
|----|---|---|
| 1. | distribution des arguments (*N0, N1, N2*) | *N2 =: Nhum* |
| 2. | distribution des prépositions locatives | *Loc N2 =: προς [pros=vers] N2* |
| 3. | interprétation des arguments | *N0 destination* |
| 4. | transformation/construction complète | *N0 V* |
| 5. | transformation/construction relative | *N1 =: Ppv* |
| 6. | complément supplémentaire | *με [me=avec] N* |
| 7. | champ lexical | *V-adj* |
| 8. | formation d'une nouvelle entrée | *από-V [apó-]* |
| *  | combinaison de No 1 et No 7 | *N0 =: V-n* |
| *  | combinaison de No 2 et No 3 | *Loc N2 =: από [apó=de] N2 source* |
| *  | combinaison de No 3 et No 7 | *V-n instrument* |

Tableau 1. L'ensemble des structures utilisées dans les tables (colonne 2) selon les différents types d'information fournis (colonne 1)

Dans ce cadre, nous avons été obligées de supprimer quelques colonnes ayant le même intitulé dans une même table. Par exemple, plusieurs colonnes intitulées *Npred* permettaient de contenir plusieurs noms prédicatifs pour chaque entrée. Nous avons regroupé dans une même case ces noms prédicatifs en les séparant par des +. À l'intérieur des tables, nous pouvons donc avoir une structure du type *x+y* où *x* et *y* sont des mots alternatifs. De plus, parmi les notations spécifiques, le grec étant une langue à cas, nous avons ajouté

---

[7] Une documentation sur l'ensemble des propriétés contenues dans la version actuelle des tables est disponible sur http://users.auth.gr/~kiroanni > Documentation des tables LG.

l'information du cas collée au nom concerné (p.ex. *N0 V κατά [katá=contre] N2humgenitif*).

Plus précisément, concernant les cas, nous n'avons pas intégré l'information du cas à tous les noms, mais seulement si les règles générales de la langue ne sont pas respectées. Comme règles générales en grec moderne, nous considérons les suivantes :

  a. toutes les prépositions introduisent un groupe nominal à l'accusatif (p.ex. *από [apó=de] N0*)
  b. le sujet de la phrase (*N0*) se trouve toujours au nominatif
  c. le complément d'objet direct (*N1*) se trouve toujours à l'accusatif

Il arrive souvent que ces règles ne soient pas respectées : il existe des prépositions (*κατά [katá=contre]*) qui exigent un groupe nominal au génitif ; un sujet de la phrase peut être introduit par une préposition dans le cadre d'une transformation, et alors il se met au cas qu'exige la préposition (ex. (1)) ; un complément d'objet direct, dans le cadre d'une transformation, peut prendre la position syntaxique d'un sujet et se mettre au nominatif (ex. (2)).

(1)   Το γεγονός ότι      μετάνιωσε      αμφισβητείται        από τον Πέτρο
                                                              [apó=par] N0accusatif
      Le fait qu'           il a regretté    est contesté         par Pierre

(2)   Η πόρτα ανοίγει (transformation de la phrase Εγώ ανοίγω την πόρτα)
      N1nominatif V
      La porte ouvre (transformation de la phrase J'ouvre la porte)

Pour traiter l'exemple (1), nous avons un ordre de priorité pour les trois règles précédentes (ordre a, b, c), ce qui permet de les appliquer successivement. Ainsi, la deuxième règle sera appliquée seulement si la première règle n'est pas valable, ce qui signifie que le sujet de la phrase se trouve au nominatif sauf s'il est précédé d'une préposition. Pour faire face aux exceptions, nous avons ajouté l'information du cas pour le nom concerné. Nous avons donc les notations *N1nominatif, κατά [katá=contre] N2humgenitif*, etc.

En plus des règles ci-dessus et à cause de la présence de verbes copules et de verbes transitifs dans les tables, nous avons précisé dans un fichier à part[8] quels sont les verbes qui sont transitifs et alors exigent un complément d'objet direct à l'accusatif, et quels sont les verbes copules qui exigent un attribut au nominatif. Ainsi, nous n'avons pas mis ces informations dans les intitulés des propriétés, car il s'agit d'une information qui ne concerne pas une construction donnée mais toute construction contenant le verbe en question. Par contre, dans quelques constructions nous avons mis l'information *datif* qui n'est pas exploitable pour le moment, mais qui est une information linguistique que l'on souhaite garder.

3. En vue de l'exploitation informatique des tables et de l'intégration d'un lexique syntaxique dans un analyseur syntaxique, il a fallu ajouter quelques informations lexicales qui étaient implicites dans les tables. Les informations lexicales ajoutées sont la forme que peut prendre le participe passé (différents cas) selon la structure ou la phrase complétive, la forme médiopassive du verbe (en grec, le verbe médiopassif est une entrée différente de la forme active, aussi bien dans les dictionnaires que dans les tables), le participe passé du verbe ainsi que l'adjectif dérivé du verbe, ayant des suffixes divers (*Sfx = τος [tos], Sfx = ος [os], Sfx = τικός [tikós]*). Cette modification concerne 9% de l'ensemble des changements dans les tables. Ces changements sont effectués soit en changeant les intitulés existants (ex. (3) et (4)) , soit

---

[8] Un fichier explicitant les informations implicites dans les tables est disponible sur
http://users.auth.gr/~kiroanni > Documentation des tables LG.

en ajoutant une nouvelle colonne avec des champs lexicaux (*VP*, *Vpp* et *"V-adj, Sfx = τος [tos]"*)

(3)     Loc N2 – Ppv → Loc N2 = Ppv =: (μου+μας+σου+σας+του+τους+της)
                    [(mou+mas+sou+sas+tou+tous+tis)=(lui+en)]
(4)     N1 = Ppv → N1 = Ppv =: (με+μας+σε+σας+τον+τους+τη+την+τις+το+τα)
                    [(me+mas+se+sas+ton+tous+ti+tin+tis+to+ta)=(le+la+les)]

De plus, étant donné le petit nombre de tables du Lexique-Grammaire pour le grec moderne, nous avons ajouté provisoirement une colonne intitulée *N0 Vsup Npred* dans 12 tables pour pouvoir exploiter les verbes supports (*Vsup*) et les noms prédicatifs (*Npred*) qui existent en tant que champs lexicaux dans les tables.

4. Quelques tables grecques du Lexique-Grammaire contenaient déjà les propriétés définitoires (Voskaki, 2011). En outre, d'autres tables étant intégrées dans une super-table (Yannacopoulou, 2005), contenaient l'ensemble des propriétés apparaissant dans les autres tables, même si elles n'étaient pas pertinentes pour la table en question. Pour chaque table, nous avons enlevé les colonnes avec les propriétés définitoires ainsi que les propriétés non pertinentes (5% des changements des tables).

5. Enfin, en modifiant les tables, nous sommes tombées sur quelques fautes linguistiques que nous avons corrigées et qui représentent moins de 1% des modifications des tables.

Regroupons les symboles ajoutés à cause des particularités de la langue grecque :

a. Traits sémantiques : *argent*, *transport*. Le trait *transport* se réfère à tous les noms qui désignent un moyen de transport (*train*, *avion*, etc.) Le trait *argent* (ex. (5)) se différencie du trait *monnaie* car il ne se réfère pas à une unité monétaire (p.ex. *francs*, *euros*) mais à tout ce qui a une valeur, qui correspond à une somme d'argent (p.ex. *subvention, bourse,* etc.)

(5)     Propriété : *N1 = : Nargent* (υποτροφία [ypotrofía=bourse]
        Entrée acceptant la propriété : *επενδύω [ependýo=investir]*

        Αυτός επένδυσε όλη την υποτροφία του σε ακίνητα
        Il a investi toute sa bourse dans l'immobilier

b. Rôle thématique : *moyen-destination* (ex. (6)). Il s'agit d'un nouveau rôle thématique qui est attribué à un complément essentiel locatif *(N2)* qui désigne à la fois le moyen et la destination.

(6)     Propriété : *Loc N2 =: (με[me=avec]+σε[se=à]) N2 moyen-destination*
        Entrée acceptant la propriété : *κρύβω [krývo=cacher]*

        Αυτός κρύβει το ψωμί με την πετσέτα / Αυτός κρύβει το ψωμί στην πετσέτα
        Il cache le pain avec la serviette / Il cache le pain à la serviette

c. Concernant les phrases complétives, nous avons utilisé la notation *Pcomp0* pour exprimer qu'il s'agit d'une phrase complétive qui a la position syntaxique d'un sujet (*N0*) (Kyriacopoulou, 2005). Cette complétive est par ailleurs décrite explicitement dans d'autres colonnes de la table, comme l'ensemble des complétives, avec une notation du type *Px* où *x* désigne la conjonction qui introduit la phrase en question (*Pνα*, *Pότι*). Ceci diffère du français, où c'est le mode de la complétive qui est indiqué : subjonctif ou indicatif (*Pind* ou *Psubj*). La raison de cette différenciation est qu'il existe plusieurs conjonctions qui exigent l'indicatif (*Pότι [óti=que]*, *Pπως*

*[pos=que]*, *Pαν [an=si]*, *Pπου [pou=que]* et *Pμήπως [mípos=si]*). Dans un fichier à part, nous avons mis toutes les conjonctions utilisées dans une phrase complétive avec leur mode exigé (voir note 7). De plus, contrairement au français, c'est la complétive (et non l'infinitive) qui peut être contrôlée par le sujet *N0* (ex. (7)).

(7)  Ο Πέτρος    αμελεί      να του τηλεφωνήσει
                             N1 =: να (na=que) V0
     Pierre      néglige     de lui téléphoner

Enfin, les complétives peuvent être de plus nominalisées, c'est-à-dire introduites par *το [to=le]* ou *το γεγονός [to gegonés=le fait]*.

d. Quant aux préfixes, ils ont une interprétation différente dans le lexique syntaxique, selon la structure de l'intitulé qui les contient (Fista, Kyriacopoulou, Martineau & Voskaki, 2008). Lorsque l'on a un intitulé contenant uniquement le préfixe avec la lettre *V* (*ξε-V [kse-]*), nous avons vu précédemment (cf. 3.1) qu'il fallait spécifier une nouvelle entrée constituée du préfixe suivi du verbe (sans tiret), qui accepte le même ensemble de propriétés et de transformations que le verbe sans préfixe. Lorsqu'ils sont utilisés dans une construction (p.ex. *N0 εκ-V [ek-] N1 Loc N2 source*), pour interpréter le prédicat, il faut ajouter le préfixe à l'entrée verbale dans cette construction uniquement. De plus, dans une construction, le préfixe peut être ajouté au participe passé du verbe (*Vpp*) avec une notation de la forme *εκ-Vpp [ek-]* (p.ex. *N1 είμαι [eímai=être] ξε-Vpp [kse-]*). Pour interpréter le prédicat dans une telle construction, il faut ajouter le préfixe au participe passé du verbe qui se trouve dans une colonne lexicale. Enfin, la notation *X-V* n'est pas exploitable, mais contient une information étymologique, indiquant que le verbe est formé d'un préfixe et d'un autre verbe.

## 4. Table des classes et lexique *LGLex*

Tout d'abord, toutes les propriétés définitoires vraies pour l'ensemble des entrées de chaque table (cf. section 2) ont été ajoutées. Au départ, la table des classes contenait 280 propriétés, incluant les différentes notations. À partir de la génération automatique de la table des classes, nous avons repéré les erreurs de notation pour les corriger directement dans les tables. La nouvelle table des classes générée contient 195 propriétés.

Ensuite, le script d'extraction des verbes a été réalisé, comme pour le français (Tolone, 2011). Il spécifie toutes les opérations liées à chaque propriété devant être effectuées pour toutes les tables. Cela nous a permis de générer, à l'aide de *LGExtract* (Constant & Tolone, 2010), une première version du lexique *LGLex* des verbes grecs[9], au format texte et XML. Ce lexique a vocation à décrire les tables avec les concepts manipulés par celles-ci, en un format directement exploitable dans les applications de TAL. L'une des utilisations informatiques possibles est la conversion en un autre format, comme cela a été fait pour le français (Tolone & Sagot, 2011). Dans sa version textuelle, une entrée de *LGLex* se présente comme suit :
− l'entrée commence par un identifiant indiquant sa catégorie, la table dont il provient et le numéro de l'entrée dans cette table (ID=catégorie_numTable_numEntrée) ;
− la section **lexical-info** indique les informations lexicales liées à l'entrée (le lemme et les prépositions associées à certains arguments) ;
− la section **args** décrit les distributions des différents arguments, avec éventuellement d'autres informations (traits sémantiques, mode et contrôle de la complétive, prépositions) ;
− la section **all-constructions** liste différentes constructions dans lesquelles l'entrée peut prendre part (soit nommées de façon complète avec tous les éléments dans l'ordre, soit des transformations à partir de construction de base) ;
− la section **example** illustre l'entrée.

---
[9] Le lexique *LGLex* des verbes grecs sera disponible ultérieurement sur http://www-igm.univ-mlv.fr/~ressourcesgrec.

Voici par exemple, le verbe *βγάζω [bgázo=sortir]* (ex. (8)) de la table 32GL qui a pour construction de base *N0 V N1 Loc N2 source Loc N3 destination*, dont l'argument *N2* est introduit par la préposition *από [apó=de]* et dont l'argument *N3* est introduit par *σε [se=à]*. Le *N0* est humain et le *N1* et *N2* concrets, le *N1* pouvant se pronominaliser en *le+la+les*. Les constructions montrent que les arguments *N2* et *N3* sont effaçables. Enfin, le verbe préfixé *ξαναβγάζω [ksanabgázo=resortir]* (ex. (9)) accepte les mêmes propriétés (voir dans l'extrait suivant le champs pfx-V) :

(8)   Έβγαλε          το γάλα         από το ψυγείο
      N0 V            N1 concret      από [apó=de] N2 source
      Il a sorti      le lait         du frigo

(9)   Ξαναέβγαλε      το γάλα         από το ψυγείο
      N0 ξανα- V      N1 concret      από [apó=de] N2 source
      Il a re-sorti   le lait         du frigo

```
ID=V_38GL_33
lexical-info=[cat="verb",verb=[lemma="βγάζω"],pfx-V=(verb="ξαναβγάζω",verb="παραβγάζω"),
        prepositions=(),locatifs=(locatif=[id="2",list=(prep="από")],
          locatif=[id="3",list=(prep="σε")])]
args=(const=[pos="0",dist=(comp=[cat="NP",hum="true",introd-prep=(),introd-loc=(),
                        origin=(orig="N0 =: Nhum")])],
      const=[pos="1",dist=(comp=[cat="NP",conc="true",introd-prep=(),introd-loc=(),
                        origin=(orig="N1 =: Nconc")])])
      const=[pos="2",dist=(comp=[cat="NP",conc="true",introd-prep=(),introd-loc=(),
                        origin=(orig="N2 =: Nconc")])],
all-constructions=[absolute=(construction="true::N0 V N1 Loc N2 source Loc N3 destination",
                 construction="o::N0 V N1 Loc N2 source (E+Loc N3 destination)",
                 construction="o::N0 V N1 (E+Loc N2 source) Loc N3 destination"),
          relative=(construction="N1 = Ppv =: (με+μας+σε+σας+τον+τους+τη+την+τις+το+τα)"]
example=[example=]
```

## 5. Conclusion

L'objectif est d'homogénéiser, corriger et compléter les données pour l'ensemble des tables du Lexique-Grammaire du grec, y compris les tables des noms prédicatifs. Une fois ces tables syntaxiques corrigées, elles seront converties au format *LGLex*, afin d'être exploitables dans des analyseurs syntaxiques.

Pour compléter les tables du lexique-grammaire des verbes grecs, il reste beaucoup à faire. En effet, il manque encore des entrées verbales à classifier (par exemple les verbes non transitifs) et la table des classes reste à coder avec les signes + et -. Les conventions de notation détaillées dans cet article seront à prendre en compte lors de la création de futures tables afin de garder l'ensemble cohérent. Puis, nous devons étendre notre travail aux noms prédicatifs, dont la modification des tables est déjà en cours. Ensuite, nous pourrons envisager la conversion du lexique *LGLex* au format Alexina, le format du lexique Le*fff* (Tolone & Sagot, 2011). Enfin, si l'on souhaite utiliser ce lexique syntaxique dans un analyseur syntaxique, il faudra adapter la méta-grammaire FRMG (Thomasset & de la Clergerie, 2005) du français à celle du grec, comme cela a été fait pour l'espagnol (Fernandez, 2010).